\def\year{2021}\relax
\documentclass[letterpaper]{article} 
\usepackage{aaai21}  
\usepackage{times}  
\usepackage{helvet} 
\usepackage{courier}  
\usepackage[hyphens]{url}  
\usepackage{graphicx} 
\usepackage{natbib}  
\usepackage{caption} 
\usepackage{subcaption}
\usepackage{amsmath}
\usepackage{amsfonts}
\usepackage{amsthm}
\usepackage{todonotes}
\usepackage[noend]{algpseudocode}
\usepackage{algorithm}
\algrenewcommand\algorithmicindent{0.8em}
\usepackage{multicol}
\usepackage{multirow}

\newtheorem{assumption}{Assumption}[section]
\newtheorem{theorem}{Theorem}[section]

\newcommand\cmax{\textsc{Cmax}}
\newcommand\cmaxpp{\textsc{Cmax++}}
\newcommand\acmaxpp{\textsc{A-Cmax++}}

\newcommand\statespace{\mathbb{S}}
\newcommand{\actionspace}{\mathbb{A}}
\newcommand{\goalspace}{\mathbb{G}}
\newcommand{\Mhat}{\hat{M}}
\newcommand\fhat{\hat{f}}
\newcommand\incorrectset{\mathcal{X}}
\newcommand\Mtilde{\tilde{M}}
\newcommand\ctilde{\tilde{c}}
\newcommand\best{\mathsf{best}}
\newcommand\Vhat{\hat{V}}
\newcommand\Vtilde{\tilde{V}}
\newcommand{\buffer}{\mathcal{D}}
\newcommand\loss{\mathcal{L}}
\newcommand\trainingset{\mathbb{X}}

\title{\cmaxpp{} : Leveraging Experience in Planning and Execution using
  Inaccurate Models}
\author {
        Anirudh Vemula\textsuperscript{\rm 1},
        J. Andrew Bagnell\textsuperscript{\rm 2},
        Maxim Likhachev\textsuperscript{\rm 1} \\
}
\affiliations {
    \textsuperscript{\rm 1} Robotics Institute, Carnegie Mellon University \\
    \textsuperscript{\rm 2} Aurora Innovation \\
    vemula@cmu.edu, dbagnell@ri.cmu.edu, maxim@cs.cmu.edu
}

\begin{document}
\maketitle

\begin{abstract}
Given access to accurate dynamical models, modern planning approaches
are effective in computing feasible and optimal plans for repetitive
robotic tasks. However, it is difficult to model the true dynamics of
the real world before execution, especially for tasks requiring
interactions with objects whose parameters are unknown. A recent
planning approach, \cmax{}, tackles this problem by adapting the
planner online during execution to bias the resulting plans away from
inaccurately modeled regions. \cmax{}, while being provably guaranteed
to reach the goal, requires strong assumptions on the accuracy of the
model used for planning and fails to improve the quality of the solution
over repetitions of the same task. In this paper we propose \cmaxpp{},
an approach that leverages real-world experience to improve the
quality of resulting plans over successive repetitions of a robotic
task.
\cmaxpp{} achieves this by integrating model-free learning
using acquired experience with model-based planning using the
potentially inaccurate model. We provide provable guarantees on the
completeness and asymptotic convergence of \cmaxpp{} to the optimal
path cost as the number of repetitions increases. \cmaxpp{} is also shown
to outperform baselines in simulated robotic tasks including 3D
mobile robot navigation where the track friction is incorrectly
modeled, and a 7D pick-and-place task where the mass of the object is
unknown leading to discrepancy between true and modeled
dynamics.\footnote{A blog post summarizing this work can be found at
\url{https://vvanirudh.github.io/blog/cmaxpp/}}
\end{abstract}

\section{Introduction}
\label{sec:introduction}

We often require robots to perform tasks that are highly repetitive,
such as picking and placing objects in assembly tasks and navigating
between locations in a warehouse. For such tasks,
robotic planning algorithms have been highly effective in cases where
system dynamics is easily specified by an efficient forward
model~\cite{DBLP:conf/icra/BerensonAG12}. However, for
tasks involving interactions with objects, dynamics are very
difficult to model without
complete knowledge of the parameters of the
objects such as mass and friction~\cite{DBLP:journals/ijrr/JiX01}. 
Using
inaccurate models for planning can result in plans
that are ineffective and fail to complete the
task~\cite{DBLP:journals/ral/McConachiePMB20}. 
In addition for such
repetitive tasks, we expect the robot's task performance to
improve, leading to efficient plans in later repetitions.
Thus, we need
a planning approach that can use potentially inaccurate models while leveraging
experience from
past executions to complete the task in each repetition, and improve
performance across repetitions.

\begin{figure}[t]
  \centering
  \begin{subfigure}{.49\columnwidth}
    \includegraphics[width=\linewidth]{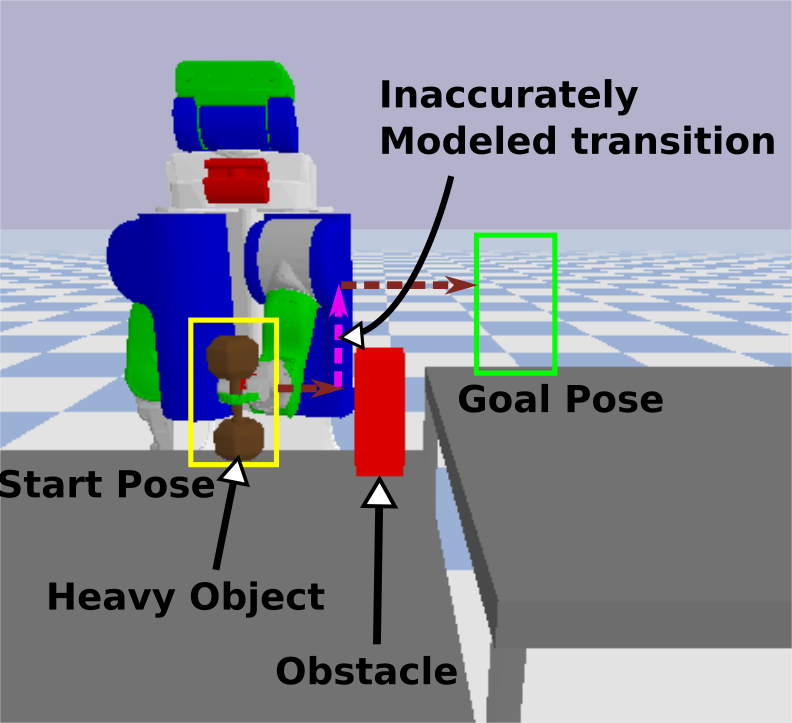}
  \end{subfigure}
  \begin{subfigure}{.49\columnwidth}
    \includegraphics[width=\linewidth]{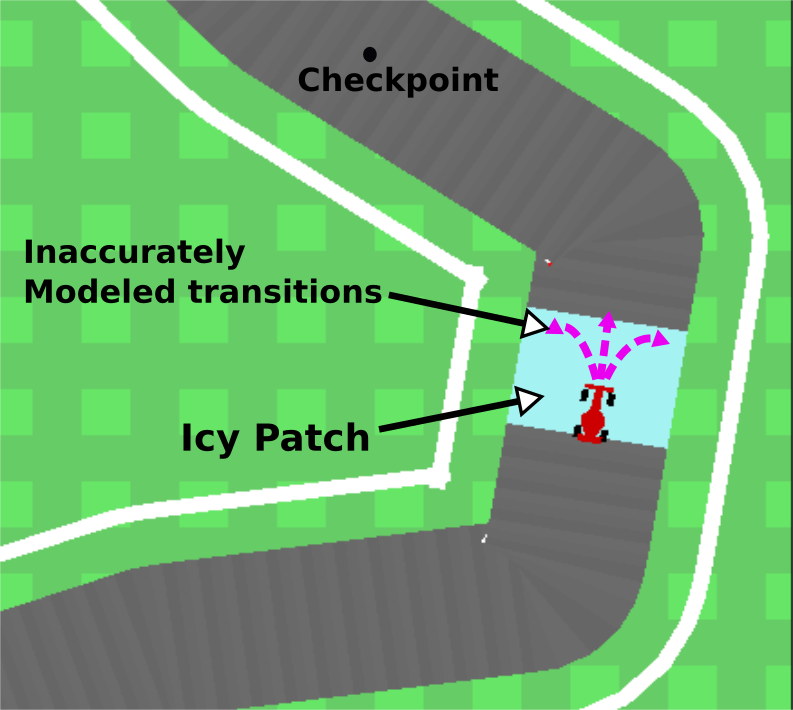}
  \end{subfigure} 
  \caption{(left) PR2 lifting a heavy dumbbell, that is modeled as
    light, to a goal location that is higher than the start location
    resulting in dynamics that are inaccurately modeled (right) Mobile robot
    navigating around a track with icy patches with unknown friction parameters
    leading to the robot skidding. In both cases, any path to the goal
    needs to contain a
    transition (pink) whose dynamics are not modeled accurately.}
  \label{fig:intro}
\end{figure}

A recent planning approach, \cmax{}, introduced
in~\cite{Vemula-RSS-20} adapts its planning strategy online to account
for any inaccuracies in the forward model without requiring any
updates to the dynamics of the model. \cmax{} achieves this online by
inflating the cost of any transition that is found to be incorrectly
modeled and replanning, thus biasing the resulting plans away from
regions where the model is inaccurate. It does so while maintaining
guarantees on completing the task, without any resets, in a finite
number of executions. However, \cmax{} requires
that there always exists a path from the current state of the robot to the goal
containing only transitions that have not yet been found to be incorrectly
modeled. This is a strong assumption on the accuracy of the model and
can often be violated, especially in the context of repetitive tasks.

For example, consider the task shown in Figure~\ref{fig:intro}(left)
where a robotic arm needs to repeatedly pick a heavy object, that is
incorrectly modeled as light, and place it on top of a taller table while avoiding
an obstacle. As the object is heavy, transitions that involve lifting
the object will have discrepancy between true and modeled
dynamics. However, any path from the start pose to the goal pose
requires lifting the object and thus, the resulting plan needs to
contain a transition that is incorrectly modeled. This violates the
aforementioned assumption of \cmax{}
and it ends up
inflating the cost of any transition that lifts the object,
resulting in plans that avoid lifting the object in future
repetitions. Thus, the quality of \cmax{} solution deteriorates
across repetitions and, in some cases, it even fails to complete the
task. Figure~\ref{fig:intro}(right) presents another example task
where a mobile robot is navigating around a track with icy patches
that have unknown friction parameters. Once the robot enters a patch,
any action executed results in the robot skidding, thus violating the
assumption of \cmax{} because
any path to the goal from current state will have inaccurately modeled
transitions. \cmax{}
ends up inflating the cost of all actions executed inside the icy
patch, leading to the robot being unable to find a path in future laps and
failing to complete the task. Thus, in both examples, we need a
planning approach that allows solutions to contain incorrectly modeled
transitions while ensuring that the robot reaches the goal.

In this paper we present \cmaxpp{}, an approach for interleaving
planning and execution that uses inaccurate models and leverages
experience from past executions to provably complete the task in each
repetition without any resets. Furthermore, it improves the quality of
solution across
repetitions. In contrast to \cmax{}, \cmaxpp{} requires weaker conditions to
ensure task completeness, and is provably guaranteed to
converge to a plan with optimal cost as the number of repetitions
increases.
The key idea behind \cmaxpp{} is to combine the conservative behavior
of \cmax{} that tries to avoid incorrectly modeled regions with
model-free Q-learning that tries to estimate and follow the optimal
cost-to-goal value function with no regard for any discrepancies
between modeled and true dynamics.
This enables \cmaxpp{} to compute plans that utilize
inaccurately modeled transitions, unlike \cmax{}. Based on this idea,
we present an algorithm
for small state
spaces, where we can do exact planning, and a practical algorithm for
large state spaces using function approximation techniques. We also propose
an adaptive version of \cmaxpp{} that intelligently switches between 
\cmax{} and \cmaxpp{} to combine the advantages of both approaches, and exhibits
goal-driven behavior in earlier repetitions and optimality in later repetitions.
The proposed algorithms are tested on simulated robotic tasks: 3D
mobile robot navigation where the track friction is incorrectly
modeled (Figure~\ref{fig:intro} right) and a 7D pick-and-place
task where the mass of the object is unknown (Figure~\ref{fig:intro} left).

\section{Related Work}
\label{sec:related-work}

A typical approach to planning in tasks with unknown parameters is to
use acquired experience from executions to update the dynamics of the
model and replan~\cite{DBLP:journals/sigart/Sutton91}. This works well
in practice for tasks where the forward model is flexible and can be
updated efficiently. However for real world tasks, the models used for
planning cannot be updated efficiently
online~\cite{DBLP:conf/iros/TodorovET12} and are often precomputed
offline using expensive
procedures~\cite{DBLP:conf/wafr/HauserBHL06}. Another line of works~\cite{DBLP:conf/iros/SaverianoYFL17,DBLP:conf/icml/AbbeelQN06}
seek to learn a residual dynamical model to account for the
inaccuracies in the initial model. However, it can take a
prohibitively large number of executions to learn the true dynamics,
especially in domains like deformable
manipulation~\cite{essahbi2012}. This precludes these approaches from
demonstrating a goal-driven behavior as we show in our experimental analysis.

Recent works such as \cmax{}~\cite{Vemula-RSS-20}
and~\cite{DBLP:journals/ral/McConachiePMB20} pursue an alternative
approach which does not require updating the dynamics of the model or
learning a residual component. These approaches exhibit goal-driven
behavior by focusing on completing the task and not on modeling the
true dynamics accurately. While \cmax{} achieves this by inflating the
cost of any transition whose dynamics are inaccurately modeled,
\cite{DBLP:journals/ral/McConachiePMB20} present an approach that
learns a binary classifier offline that is used online to predict
whether a transition is accurately modeled or not. Although these
methods work well in practice for goal-oriented tasks, they do not
leverage experience acquired online to improve the quality of solution
when used for repetitive tasks.


Our work is closely related to approaches that integrate
model-based planning with model-free
learning. \cite{DBLP:journals/corr/abs-2005-10872} use model-based
planning in regions where the dynamics are accurately modeled and
switch to a model-free policy in regions with high
uncertainty. However, they mostly focus on perception uncertainty and
require a coarse estimate of the uncertain region prior to execution,
which is often not available for tasks with other modalities of
uncertainty like unknown inertial parameters. A very recent work
by~\cite{lagrassa2020learning} uses a
model-based planner until a model inaccuracy is detected and switches
to a model-free policy to complete the task. Similar to our approach,
they deal with general modeling errors but rely on expert
demonstrations to learn the model-free policy. In contrast, our
approach does not require any expert demonstrations and only uses the
experience acquired online to obtain model-free value estimates
that are used within planning.

Finally, our approach is also related to the field of real-time
heuristic search which tackles the problem of efficient planning in
large state spaces with bounded planning time. In this work, we
introduce a novel planner that is inspired by
LRTA*~\cite{DBLP:journals/ai/Korf90} which limits the number of
expansions in the search procedure and interleaves execution with
planning. Crucially, our planner also interleaves planning and
execution but unlike these approaches, employs model-free value
estimates obtained from past experience within the search.





\section{Problem Setup}
\label{sec:problem-setup}

Following the notation of \cite{Vemula-RSS-20}, we consider the
deterministic shortest path problem that can be represented using  the
tuple $M = (\statespace, \actionspace, \goalspace, f, c)$ where
$\statespace$ is the state space, $\actionspace$ is the action space,
$\goalspace \subseteq \statespace$ is the non-empty set of goals,
$f:\statespace \times \actionspace \rightarrow \statespace$ is a
deterministic dynamics function, and $c:\statespace \times
\actionspace \rightarrow [0, 1]$ is the cost function. Note that we
assume that the costs lie between $0$ and $1$ but any bounded cost
function can be scaled to satisfy this assumption. Crucially, our
approach assumes that the action space $\actionspace$ is discrete, and any
goal state $g \in \goalspace$ is a cost-free termination state. The
objective of the shortest path problem is to find the least-cost path
from a given start state $s_1 \in \statespace$ to any goal state $g
\in \goalspace$ in $M$. As is typical in shortest path problems, we
assume that there exists at least one path from each state $s \in
\statespace$ to one of the goal states, and that the cost of any
transition from a non-goal state is
positive~\cite{DBLP:books/lib/Bertsekas05}. We will use $V(s)$ to
denote the state value function (a running estimate of cost-to-goal
from state $s$,) and $Q(s, a)$ to denote the
state-action value function (a running estimate of the sum of
transition cost and cost-to-goal from successor state,) for any state $s$ and action
$a$. Similarly, we will use the notation $V^*(s)$ and $Q^*(s, a)$ to
denote the corresponding optimal value functions. A value estimate is
called admissible if it underestimates the optimal value function at
all states and actions, and is called consistent if it satisfies the
triangle inequality, i.e. $V(s) \leq c(s, a) + V(f(s, a))$ and $Q(s,
a) \leq c(s, a) + V(f(s, a))$ for all $s, a$, and $V(g) = 0$ for all $g \in \goalspace$.

In this work, we focus on repetitive robotic tasks where the true
deterministic dynamics $f$ are unknown but we have access to an
approximate model described using $\Mhat = (\statespace, \actionspace,
\goalspace, \fhat, c)$ where $\fhat$ approximates the true dynamics.
In each repetition of the task, the robot acts in the environment $M$
to acquire experience over a single trajectory and reach the goal,
without access to any resets. This rules out any episodic
approach.
Since the true dynamics are unknown and can only be
discovered through executions, we consider the online real-time
planning setting where the robot has to interleave planning and
execution.
In our motivating navigation example (Figure~\ref{fig:intro} right,)
the approximate model $\hat{M}$ represents a track with no icy patches
whereas the environment $M$ contains icy patches. Thus, there is a 
discrepancy between the modeled dynamics $\fhat$ and true
dynamics $f$. Following \cite{Vemula-RSS-20}, we will refer to
state-action pairs that have inaccurately modeled dynamics as
``incorrect" transitions, and use the notation $\incorrectset
\subseteq \statespace \times \actionspace$ to denote the set of discovered
incorrect transitions. The objective in our work is for the robot to
reach a goal in each repetition, despite using an inaccurate model for
planning while improving performance, measured using the cost of
executions, across repetitions.

\section{Approach}
\label{sec:approach}

In this section, we will describe the proposed approach \cmaxpp{}. 
First, we will present a novel planner used in \cmaxpp{} that
can exploit incorrect transitions using their model-free $Q$-value
estimates. Second, we present \cmaxpp{} and its adaptive version
for small state spaces, and establish their guarantees. Finally, we
describe a practical instantiation of \cmaxpp{} for large state spaces
leveraging function approximation techniques.


\subsection{Hybrid Limited-Expansion Search Planner}
\label{sec:integrating-q-values}

During online execution, we want the robot to acquire experience
and leverage it to compute better plans. This requires a hybrid planner that
is able to incorporate value estimates obtained using past experience
in addition to model-based planning, and quickly compute the next
action to execute. To achieve this, we propose a real-time
heuristic search-based planner that performs a bounded number of
expansions and is able to utilize $Q$-value estimates for incorrect
transitions.

\begin{algorithm}[t]
  \caption{Hybrid Limited-Expansion Search}
  {\normalsize
  \begin{algorithmic}[1]
    \Procedure{$\mathtt{SEARCH}$}{$s,
      \Mhat, V, Q, \incorrectset, K$}
      \State Initialize $g(s)=0$, min-priority open list $O$, and closed list $C$
      \State Add $s$ to open list $O$ with priority $p(s) = g(s) + V(s)$
      \For{$i = 1, 2, \cdots, K$}
      	\State Pop $s_i$ from $O$
      	\If{$s_i$ is a dummy state or $s_i \in \goalspace$}
      	\State Set $s_\best \leftarrow s_i$ and go to Line~\ref{line:update}\label{line:jump}
      	\EndIf
      	\For{$a \in \actionspace$}\Comment{\textit{Expanding state $s_i$}}
      		\If{$(s_i, a) \in \incorrectset$}\Comment{\textit{Incorrect transition}}
      		\State Add a dummy state $s'$ to $O$ with priority $p(s') = g(s_i) + Q(s_i, a)$\label{line:dummy}
     		\State \textbf{continue}
      		\EndIf
      		\State Get successor $s' = \fhat(s_i, a)$
      		\State If $s' \in C$, \textbf{continue}
      		\If{$s' \in O$ and $g(s') > g(s_i) + c(s_i, a)$}
      			\State Set $g(s') = g(s_i) + c(s_i, a)$ and recompute $p(s')$
      			\State Reorder open list $O$
      		\ElsIf{$s' \notin O$}
      			\State Set $g(s') = g(s_i) + c(s_i, a)$
      			\State Add $s'$ to $O$ with priority $p(s') = g(s') + V(s')$
      		\EndIf
      	\EndFor
      	\State Add $s_i$ to closed list $C$
      \EndFor
      \State Pop $s_\best$ from open list $O$\label{line:best}
      \For{$s' \in C$}\label{line:update}
      	\State Update $V(s') \leftarrow p(s_\best) - g(s')$\label{line:update2}
      \EndFor
      \State Backtrack from $s_\best$ to $s$, and set $a_\best$ as the first action on path from $s$ to $s_\best$ in the search tree\label{line:best-action}
		      
      \Return{$a_\best$}
    \EndProcedure
  \end{algorithmic}}
  \label{alg:hybrid-limited-expansion-search}
\end{algorithm}

The planner is presented in
Algorithm~\ref{alg:hybrid-limited-expansion-search}. Given the current
state $s$, the planner constructs a lookahead search tree using at
most $K$ state expansions. For each expanded state $s_i$, if any
outgoing transition has been flagged as incorrect based on experience, i.e. $(s_i, a) \in \incorrectset$,
then the planner creates a dummy state with priority computed using
the model-free $Q$-value estimate of that transition
(Line~\ref{line:dummy}). Note that we create a dummy state because the
model $\Mhat$ does not know the true successor of an incorrect transition.
For
the transitions that are correct, we obtain successor states
using the approximate model $\Mhat$. This ensures that we rely on the
inaccurate model only for transitions that are not known to be
incorrect.
At any stage, if a dummy state is expanded then we need to terminate
the search as the model $\Mhat$ does not know any of its successors,
in which case we set the best state $s_\best$ as the dummy state (Line~\ref{line:jump}).
Otherwise, we choose $s_\best$ as the best state (lowest priority)
among the
leaves of the search tree after $K$ expansions (Line~\ref{line:best}). Finally, the best
action to execute at the current state $s$ is computed as the first
action along the path from $s$ to $s_\best$ in the search tree (Line~\ref{line:best-action}). The
planner also updates state value estimates $V$ of all expanded states
using the priority of the best state $p(s_\best)$ to make the
estimates more accurate (Lines~\ref{line:update}
and~\ref{line:update2}) similar to RTAA*~\cite{DBLP:conf/atal/KoenigL06}.

The ability of our planner to exploit incorrect transitions using
their model-free $Q$-value estimates, obtained from past experience,
distinguishes it from
real-time search-based planners such as
LRTA*~\cite{DBLP:journals/ai/Korf90} which cannot utilize model-free
value estimates during planning.
This enables \cmaxpp{} to
result in plans that utilize incorrect transitions if they enable the
robot to get to the goal with lower cost.

\subsection{\cmaxpp{} in  Small State Spaces}
\label{sec:warm-up:-small}

\cmaxpp{} in small state spaces is simple and easy-to-implement as it
is feasible to maintain value estimates in a table for all states and
actions and to explicitly maintain a running set of incorrect
transitions with fast lookup without resorting to function
approximation techniques.

\begin{algorithm}[t]
	\caption{\cmaxpp{} and \textcolor{blue}{\acmaxpp{}} in small state spaces}
	\begin{algorithmic}[1]
		\Require{Model $\Mhat$, start state $s$, initial value estimates $V$, $Q$, number of expansions $K$, $t\leftarrow 1$, incorrect set $\incorrectset \leftarrow \{\}$, Number of repetitions $N$, \textcolor{blue}{Sequence $\{\alpha_i \geq 1\}_{i=1}^N$, initial penalized value estimates $\Vtilde = V$, penalized model $\Mtilde \leftarrow \Mhat$}}
		\For{each repetition $i=1,\cdots, N$}
		\State $t \leftarrow 1$, $s_1 \leftarrow s$
		\While{$s_t \notin \goalspace$}
			\State Compute $a_t = \mathtt{SEARCH}(s_t, \Mhat, V, Q, \incorrectset, K)$ \label{line:cmaxpp-action}
			\textcolor{blue}{
			\State Compute $\tilde{a}_t = \mathtt{SEARCH}(s_t, \Mtilde, \Vtilde, Q, \{\}, K)$\label{line:cmax-action}
			\State If $\Vtilde(s_t) \leq \alpha_i V(s_t)$, assign $a_t = \tilde{a}_t$\label{line:switch}
			}
			\State Execute $a_t$ in environment to get $s_{t+1} = f(s_t, a_t)$
			\If{$s_{t+1} \neq \fhat(s_t, a_t)$}
				\State Add $(s_t, a_t)$ to the set: $\incorrectset \leftarrow \incorrectset \cup \{(s_t, a_t)\}$
				\State Update: $Q(s_t, a_t) = c(s_t, a_t) + V(s_{t+1})$
				\textcolor{blue}{
				\State Update penalized model $\Mtilde \leftarrow \Mtilde_{\incorrectset}$
				}
			\EndIf
			\State $t \leftarrow t + 1$
		\EndWhile
		\EndFor
	\end{algorithmic}
	\label{alg:small-state-spaces}
\end{algorithm}

The algorithm is presented in Algorithm~\ref{alg:small-state-spaces}
(only the text in black.) \cmaxpp{} maintains a running estimate of the set
of incorrect transitions $\incorrectset$, and updates the set
whenever it encounters an incorrect state-action pair during
execution. Crucially, unlike \cmax{}, it maintains a $Q$-value estimate
for the incorrect transition that is used during planning in
Algorithm~\ref{alg:hybrid-limited-expansion-search}, thereby enabling
the planner to compute paths that contain incorrect transitions. It is
also important to note that, like \cmax{}, \cmaxpp{} never updates the
dynamics of the model. However, instead of
using the penalized model for planning as \cmax{} does, \cmaxpp{} uses the initial
model $\Mhat$, and utilizes both model-based planning and model-free
$Q$-value estimates to replan a path from the current state to a goal.

The downside of \cmaxpp{} is that estimating $Q$-values from online
executions can be inefficient as it might take many executions before
we obtain an accurate $Q$-value estimate for an incorrect
transition. This has been extensively studied in the past and is a
major disadvantage of model-free
methods~\cite{DBLP:conf/colt/SunJKA019}. As a result of this
inefficiency, \cmaxpp{} lacks the goal-driven behavior of \cmax{} in
early repetitions of the task, despite achieving optimal behavior in
later repetitions. In the next section, we present an adaptive version
of \cmaxpp{} (\acmaxpp{}) that combines the goal-driven behavior of \cmax{} with
the optimality of \cmaxpp{}.

\subsection{Adaptive Version of \cmaxpp{}}
\label{sec:adaptive}


\subsubsection{Background on \cmax{}}
\label{sec:background-cmax}
Before we describe \acmaxpp{}, we will start by summarizing
\cmax{}. For more details, refer 
to~\cite{Vemula-RSS-20}.
At each time step $t$ during execution, \cmax{} maintains a running
estimate of the incorrect set $\incorrectset$, and constructs a
penalized model specified by the tuple $\Mtilde_\incorrectset =
(\statespace, \actionspace, \goalspace, \fhat,
\ctilde_{\incorrectset})$ where the cost function
$\ctilde_{\incorrectset}(s, a) = |\statespace|$ if $(s, a) \in
\incorrectset$, else $\ctilde_{\incorrectset}(s, a) = c(s, a)$. In
other words, the cost of any transition found to be incorrect is
set high (or inflated) while the cost of other transitions are the
same as in $\Mhat$.
\cmax{} uses the penalized model $\Mtilde_{\incorrectset}$ to plan a path from the
current state $s_t$ to a goal state. Subsequently, \cmax{} executes
the first action $a_t$ along the path and observes if the true
dynamics and model dynamics differ on the executed action. If so, the
state-action pair $(s_t, a_t)$ is appended to the incorrect set
$\incorrectset$  and the penalized model $\Mtilde_{\incorrectset}$ is updated.
\cmax{} continues to do this at every
timestep until the robot reaches a goal state.

Observe that the inflation of cost for any incorrect state-action pair
biases the planner to ``explore" all other state-action pairs that are
not yet known to be incorrect before it plans a path using an
incorrect transition. This induces a goal-driven behavior in the
computed plan that enables \cmax{} to quickly find an alternative path
and not waste executions learning the true dynamics

\subsubsection{\acmaxpp{}}
\label{sec:acmaxpp}
\acmaxpp{} is presented in
Algorithm~\ref{alg:small-state-spaces} (black and blue text.)
\acmaxpp{} maintains a running estimate of incorrect set
$\incorrectset$ and constructs the penalized model $\Mtilde$ at each
time step $t$, similar to \cmax{}. For any state at time step $t$, we
first compute the best action $a_t$ based on the approximate model
$\Mhat$ and the model-free $Q$-value estimates
(Line~\ref{line:cmaxpp-action}.) In addition, we also compute the best
action $\tilde{a}_t$ using the penalized model $\Mtilde$, similar to
\cmax{}, that inflates the cost of any incorrect transition
(Line~\ref{line:cmax-action}.) The crucial step in \acmaxpp{}
is Line~\ref{line:switch} where we compare the penalized value
$\Vtilde(s_t)$ (obtained using penalized model $\Mtilde$) and the
non-penalized value $V(s_t)$ (obtained using approximate model $\Mhat$
and $Q$-value estimates.) Given a sequence $\{\alpha_i \geq 1\}$ for
repetitions $i=1,\cdots,N$ of the task, if $\Vtilde(s_t) \leq
\alpha_iV(s_t)$, then we execute action $\tilde{a}_t$, else we execute
$a_t$. This implies that if the cost incurred by following \cmax{}
actions in the future is within $\alpha_i$ times the cost incurred by
following \cmaxpp{} actions, then we prefer to execute \cmax{}.

If
the sequence $\{\alpha_i\}$ is chosen to be non-increasing such that
$\alpha_1 \geq \alpha_2 \cdots \geq \alpha_N \geq 1$, then we can observe that
\acmaxpp{} has the desired anytime-like behavior. It remains goal-driven in
early repetitions, by choosing \cmax{} actions, and
converges to optimal behavior in later repetitions, by choosing
\cmaxpp{} actions. Further, the
executions needed to obtain accurate $Q$-value estimates is
distributed across repetitions ensuring that \acmaxpp{} does
not have poor performance in any single repetition. Thus,
\acmaxpp{} combines the advantages of both \cmax{} and \cmaxpp{}.

\subsection{Theoretical Guarantees}
\label{sec:guarantees}

We will start with formally stating the assumption needed by
\cmax{} to ensure completeness:
\begin{assumption}[\cite{Vemula-RSS-20}]
	Given a penalized model $\Mtilde_{\incorrectset_t}$ and the
        current state $s_t$ at any time step $t$, there always exists
        at least one path from $s_t$ to a goal that does not contain
        any state-action pairs $(s, a)$ that are known to be
        incorrect, i.e. $(s, a) \in \incorrectset_t$.  
        \label{assumption:cmax}      
\end{assumption}
Observe that the above assumption needs to be valid at every time step
$t$ before the robot reaches a goal and thus, can be hard to satisfy.
Before we state the theoretical guarantees for \cmaxpp{}, we
need the following assumption on the approximate model $\Mhat$ that is
used for planning:
\begin{assumption}
	The optimal value function $\hat{V}^*$ using the dynamics of
        approximate model $\Mhat$ underestimates the optimal value
        function $V^*$ using the true dynamics of $M$ at all states,
        i.e. $\Vhat^*(s) \leq V^*(s)$ for all $s \in
        \statespace$.    
	\label{assumption:cmaxpp}
\end{assumption}
In other words, if there exists a path from any state $s$ to a goal
state in the environment $M$, then there exists a path with the same
or lower cost from $s$ to a goal in the approximate model $\Mhat$. In
our motivating example of pick-and-place (Figure~\ref{fig:intro}
left,) this assumption is satisfied if the object is modeled as light
in $\Mhat$, as the object being heavy in reality can only increase the cost. 
This assumption was also considered in previous
works such as~\cite{DBLP:conf/aaai/Jiang18} and is known as the
\textit{Optimistic Model Assumption}.  


We can now state the following guarantees:
\begin{theorem}[Completeness]
	Assume the initial value estimates $V, Q$ are admissible and consistent. Then we have,
	\begin{enumerate}
		\item If Assumption~\ref{assumption:cmaxpp} holds then using either \cmaxpp{} or \acmaxpp{}, the robot is guaranteed to reach a goal state in at most $|\statespace|^3$ time steps in each repetition.
		\item If Assumption~\ref{assumption:cmax} holds then
                  (a) using \acmaxpp{} with a large enough $\alpha_i$
                  in any repetition $i$ (typically true for early
                  repetitions,) the robot is guaranteed to reach a
                  goal state in at most $|\statespace|^2$ time steps,
                  and (b) using \cmaxpp{}, it is guaranteed to reach a goal state in at most $|\statespace|^3$ time steps in each repetition
	\end{enumerate}       
	\label{theorem:completeness}
      \end{theorem}
      \noindent\textit{Proof Sketch.}
	The first part of theorem follows from the analysis of
        Q-learning for systems with deterministic
        dynamics~\cite{DBLP:conf/aaai/KoenigS93}. In the worst case,
        if the model is incorrect everywhere and if
        Assumption~\ref{assumption:cmaxpp} (or Assumption~\ref{assumption:cmax}) holds then,
        Algorithm~\ref{alg:small-state-spaces} reduces to Q-learning,
        and hence we can borrow its worst case bounds. The second part
        of the theorem concerning \acmaxpp{} follows from the completeness proof of
        \cmax{}. $\square$
\begin{theorem}[Asymptotic Convergence]
	Assume Assumption~\ref{assumption:cmaxpp} holds, and that the
        initial value estimates $V, Q$ are admissible and
        consistent. For sufficiently large number of repetitions $N$,
        there exists an integer $j \leq N$ such that the robot follows
        a path with the optimal cost to the goal using \cmaxpp{} in
        Algorithm~\ref{alg:small-state-spaces} in repetitions $i \geq
        j$.        
	\label{theorem:convergence}
\end{theorem}
\noindent\textit{Proof Sketch.}
	The guarantee follows from the asymptotic convergence of
        Q-learning~\cite{DBLP:conf/aaai/KoenigS93}.
        $\square$      

It is important to note that
the conditions required for Theorem~\ref{theorem:completeness}
are weaker than the conditions required for completeness of
\cmax{}. Firstly,
if either Assumption~\ref{assumption:cmax} or
Assumption~\ref{assumption:cmaxpp} holds then \cmaxpp{} can be shown
to be complete, but \cmax{} is guaranteed to be complete only under
Assumption~\ref{assumption:cmax}.
Furthermore,
Assumption~\ref{assumption:cmaxpp} only needs to hold
for the approximate model $\Mhat$ we start with, whereas
Assumption~\ref{assumption:cmax} needs to be satisfied for every penalized model $\Mtilde$ constructed at any time
step $t$ during execution.


\subsection{Large State Spaces}
\label{sec:large-state-spaces}

\begin{algorithm}[t]
	\caption{\cmaxpp{} in large state spaces}
	\begin{algorithmic}[1]
		\Require{Model $\Mhat$, start state $s$, value function approximators $V_\theta$, $Q_\zeta$, number of expansions $K$, $t\leftarrow 1$, Discrepancy threshold $\xi$, Radius of hypersphere $\delta$, Set of hyperspheres $\incorrectset^\xi \leftarrow \{\}$, Number of repetitions $N$, Batch size $B$, State buffer $\buffer_S$, Transition buffer $\buffer_{SA}$, Learning rate $\eta$, Number of updates $U$}
		\For{each repetition $i=1,\cdots, N$}
		\State $t \leftarrow 1$, $s_1 \leftarrow s$
		\While{$s_t \notin \goalspace$}
			\State Compute $a_t = \mathtt{SEARCH}(s_t, \Mhat, V_\theta, Q_\zeta, \incorrectset^\xi, K)$
			\State Execute $a_t$ in environment to get $s_{t+1} = f(s_t, a_t)$
			\If{$d(s_{t+1}, \fhat(s_t, a_t)) > \xi$}
				\State Add hypersphere: $\incorrectset^\xi \leftarrow \incorrectset^\xi \cup \{\mathsf{sphere}(s_t, a_t, \delta)\}$\label{line:hypersphere}
			\EndIf
			\State Add $s_t$ to $\buffer_S$, and $(s_t, a_t, s_{t+1})$ to $\buffer_{SA}$
			\For{$u=1,\cdots, U$}\Comment{\textit{Approximator updates}}\label{line:approximation_updates}
			\State $\mathtt{Q\_UPDATE}(Q_\zeta, V_\theta, \buffer_{SA})$
			\State $\mathtt{V\_UPDATE}(V_\theta, Q_\zeta, \buffer_S, \incorrectset^\xi)$
			\EndFor
			\State $t \leftarrow t + 1$
		\EndWhile
		\EndFor
	\Procedure{$\mathtt{Q\_UPDATE}$}{$Q_\zeta, V_\theta, \buffer_{SA}$}\label{line:q-update}
		\State Sample $B$ transitions from $\buffer_{SA}$ with replacement\label{line:transition-sample}
		\State Construct training set $\trainingset_Q = \{((s_i, a_i), Q(s_i, a_i))\}$ for each sampled transition $(s_i, a_i, s_i')$ and compute $Q(s_i, a_i) = c(s_i, a_i) + V_\theta(s_i')$
		\State Update: $\zeta \leftarrow \zeta - \eta \nabla_\zeta \loss_Q(Q_\zeta, \trainingset_Q)$
	\EndProcedure
	\Procedure{$\mathtt{V\_UPDATE}$}{$V_\theta, Q_\zeta, \buffer_S, \incorrectset^\xi$}\label{line:v-update}
		\State Sample $B$ states from $\buffer_S$ with replacement\label{line:state-sample}
		\State Call $\mathtt{SEARCH}(s_i, \Mhat, V_\theta, Q_\zeta, \incorrectset^\xi, K)$ for each sampled $s_i$ to get all states on closed list $s_i'$ and their corresponding value updates $V(s_i')$ to construct training set $\trainingset_V = \{(s_i', V(s_i')\}$
		\State Update: $\theta \leftarrow \theta - \eta \nabla_\theta \loss_V(V_\theta, \trainingset_V)$
	\EndProcedure
	\end{algorithmic}
	\label{alg:large-state-spaces}
\end{algorithm}

In this section, we present a practical instantiation of \cmaxpp{} for
large state spaces where it is infeasible to maintain tabular value
estimates and the incorrect set $\incorrectset$ explicitly. Thus, we
leverage function approximation techniques to maintain these
estimates. Assume that there exists a 
metric $d$ under which $\statespace$ is bounded. We relax the definition
of incorrect set using this metric to define $\incorrectset^\xi$ as
the set of all $(s, a)$ pairs such that $d(f(s, a), \fhat(s, a)) >
\xi$ where $\xi \geq 0$. Typically, we chose $\xi$ to allow for small
modeling discrepancies that can be compensated by a low-level path
following controller.

\cmaxpp{} in large state spaces is presented in
Algorithm~\ref{alg:large-state-spaces}. The algorithm closely follows
\cmax{} for large state spaces presented in~\cite{Vemula-RSS-20}. The
incorrect set $\incorrectset^\xi$ is maintained using sets of
hyperspheres with each set corresponding to a discrete
action. Whenever the agent executes an incorrect state-action $(s,
a)$, \cmaxpp{} adds a hypersphere centered at $s$ with radius
$\delta$, as measured using metric $d$, to the incorrect set
corresponding to action $a$. In future
planning, any state-action pair $(s', a')$ is declared incorrect if
$s'$ lies inside any of the hyperspheres in the incorrect set
corresponding to action $a'$.
After each execution, \cmaxpp{} proceeds to update the value
function approximators (Line~\ref{line:approximation_updates}) by
sampling previously executed transitions and visited states from
buffers and performing gradient descent steps
(Procedures~\ref{line:q-update} and~\ref{line:v-update}) using mean
squared loss
functions given by $\loss_Q(Q_\zeta, \trainingset_Q) =
                                     \frac{1}{2|\trainingset_Q|}\sum_{(s_i,
                                     a_i) \in \trainingset_Q} (Q(s_i,
                                     a_i) - Q_\zeta(s_i, a_i))^2$ and $\loss_V(V_\theta, \trainingset_V) = \frac{1}{2|\trainingset_V|}\sum_{s_i \in \trainingset_V} (V(s_i) - V_\theta(s_i))^2$.

By using hyperspheres, \cmaxpp{} ``covers" the set of incorrect
transitions, and enables fast lookup using KD-Trees in the state
space. Like
Algorithm~\ref{alg:small-state-spaces}, we never update the
approximate model $\Mhat$ used for planning. However, unlike
Algorithm~\ref{alg:small-state-spaces}, we update the value estimates
for sampled previous transitions and states
(Lines~\ref{line:transition-sample} and~\ref{line:state-sample}). This ensures that the
global function approximations used to maintain value estimates
$V_\theta, Q_\zeta$ have good generalization beyond the current state
and action. Algorithm~\ref{alg:large-state-spaces} can also be extended in a similar fashion as Algorithm~\ref{alg:small-state-spaces} to include \acmaxpp{} by maintaining a penalized value function approximation and updating it using gradient descent.

\section{Experiments}
\label{sec:experiments}

We test the efficiency of \cmaxpp{} and \acmaxpp{} on
simulated robotic tasks emphasizing their performance in each
repetition of the task, and improvement across
repetitions\footnote{The code to reproduce our experiments can be 
found at \url{https://github.com/vvanirudh/CMAXPP}.}. In each
task, we start the next repetition only if the robot reached a goal in
previous repetition.

\subsection{3D Mobile Robot Navigation with Icy Patches}
\label{sec:simulated-3d-mobile}

In this experiment, the task is for a mobile robot with Reed-Shepp
dynamics~\cite{reeds1990} to navigate around a track $M$ with icy
patches (Figure~\ref{fig:intro} right.) This
can be represented as a planning problem in 3D discrete state space
$\statespace$ with any state represented using the tuple $(x, y,
\theta)$ where $(x, y)$ is the 2D position of the robot and $\theta$
describes its heading. The XY-space is discretized into $100 \times
100$ grid and the $\theta$ dimension is discretized into $16$
cells. We construct a lattice
graph~\cite{DBLP:journals/jfr/PivtoraikoKK09} using $66$ motion
primitives that are pre-computed offline respecting the differential
constraints on the motion of the robot. The model $\Mhat$ used for
planning contains the same track as $M$ but without any icy patches,
thus the robot discovers transitions affected by icy patches only
through executions.

\begin{figure}[t]
  \centering
  \includegraphics[width=\linewidth]{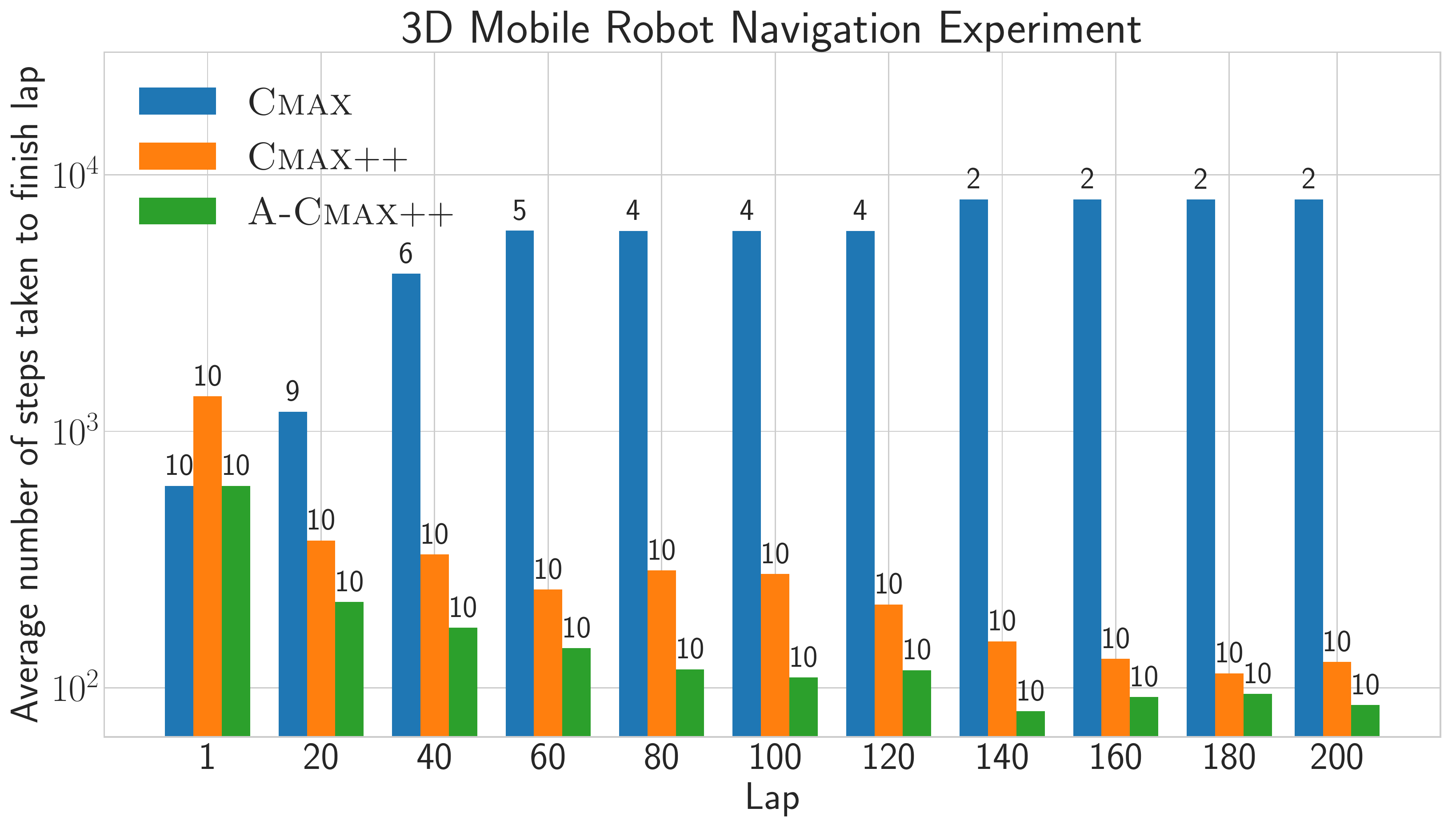}
  \caption{Number of steps taken to finish a lap averaged across 10 instances each with $5$ icy patches placed randomly around the track. The number above each bar reports the number of instances in which the robot was successful in finishing the respective lap within $10000$ time steps.}
  \label{fig:car_racing}
\end{figure}

Since the state space is small, we use
Algorithm~\ref{alg:small-state-spaces} for \cmaxpp{} and
\acmaxpp{}. For \acmaxpp{}, we use a non-increasing sequence with
$\alpha_i = 1 + \beta_i$ where $\beta_1 = 100$ and $\beta_i$ is
decreased by 2.5 after every $5$ repetitions (See Appendix for more details on choosing the sequence.) We
compare both algorithms with \cmax{}. For all the approaches, we
perform $K=100$ expansions. Since
the motion primitives are computed offline using an expensive
procedure, it is not feasible to update the dynamics of model $\Mhat$
online and hence, we do not compare with any model learning baselines. We
also conducted several experiments with model-free Q-learning, and found that it
performed poorly requiring a very large number of executions and finishing
only 10 laps in the best case. Hence, we do not include it in our
results shown in Figure~\ref{fig:car_racing}.

\cmax{} performs
well in the early laps computing paths with lower costs compared to
\cmaxpp{}. However, after a few laps the robot using \cmax{} gets
stuck within an icy patch and does not make any more progress. Observe
that when the robot is inside the icy patch,
Assumption~\ref{assumption:cmax} is violated and \cmax{} ends up
inflating all transitions that take the robot out of the patch leading
to the robot finishing $200$ laps in $2$ out of $10$
instances. \cmaxpp{}, on the other hand, is suboptimal in the initial
laps, but converges to paths with lower costs in later
laps. More importantly, the robot using \cmaxpp{} manages to finish
$200$ laps in all $10$ instances. \acmaxpp{} also successfully
finishes $200$ laps in all $10$ instances. However, it 
outperforms both \cmax{} and \cmaxpp{} in all laps by intelligently switching between them achieving goal-driven behavior in early laps and optimal behavior in later laps.
Thus, \acmaxpp{} combines the
advantages of \cmax{} and \cmaxpp{}.

\subsection{7D Pick-and-Place with a Heavy Object}
\label{sec:simulated-7d-pick}

\begin{table*}[t]
  \centering
  \small
  \setlength\tabcolsep{4pt}
  \begin{tabular}{|c|c|c|c|c|c|c|c|c|c|c|}
    \hline
    \textit{Repetition$\rightarrow$} & \multicolumn{2}{c|}{${1}$} &
                                                                  \multicolumn{2}{c|}{${5}$} & \multicolumn{2}{c|}{${10}$} & \multicolumn{2}{c|}{${15}$} & \multicolumn{2}{c|}{${20}$} \\
    \cline{2-11}
    & \textit{Steps} & \textit{Success} & \textit{Steps} & \textit{Success} & \textit{Steps} & \textit{Success} &
                                                                     \textit{Steps}
                         & \textit{Success} &
                                                       \textit{Steps} & \textit{Success} \\
    \hline
    \textbf{\cmax{}} & $\mathbf{17.8 \pm 3.4}$ & $100\%$& $13.6 \pm 0.5$ & $60\%$ & $18 \pm
                                                     0$ & $20\%$ & $15
                                                                   \pm
                                                                   0$
                         & $20\%$ &
    $15 \pm 0$ & $20\%$\\
    \hline
    \textbf{\cmaxpp{}} & $\mathbf{17 \pm 4.9}$ & $100\%$ & $14.2 \pm 3.3$ &
                                                                   $100\%$
                                                                                                                                                   & $\mathbf{10.6 \pm 0.3}$ & $100\%$
                                  & $\mathbf{11 \pm 0}$ &  $100\%$ & $\mathbf{10.8 \pm
                                                            0.1}$ & $100\%$ \\
    \hline
    \textbf{\acmaxpp{}} & $\mathbf{17.8 \pm 3.4}$ & $100\%$ & $\mathbf{11.6 \pm 0.7}$ &
                                                                      $100\%$
                                                                                                                                                   & $17 \pm 6$ & $100\%$
                                  & $\mathbf{10.4 \pm 0.3}$ & $100\%$ & $\mathbf{10.6
                                                               \pm
                                                               0.4}$ & $100\%$ \\
    \hline
    \textbf{Model KNN} & $40.6 \pm 7.3$ & $100\%$ & $12.8 \pm 1.3$ & $100\%$ & $29.6 \pm
                                                      16.1$ & $100\%$ &
                                                                   $15.8\pm
                                                                   2.9$
                                                                        &
                                                                          $100\%$
                                         & $12.4 \pm 1.4$ & $100\%$\\
    \hline
    \textbf{Model NN} & $56 \pm 16.2$ & $100\%$& $208.2 \pm 92.1$ & $80\%$ & $124.5 \pm
                                                     81.6$ & $40\%$ & $28
                                                                  \pm
                                                                  7.7$
                                                                      &
                                                                        $40\%$
                                         & $37.5 \pm 20.1$ & $40\%$\\
    \hline
    \textbf{Q-learning} & $172.4\pm 75$ & $100\%$& $23.2\pm 10.3$ & $80\%$ & $26.5 \pm
                                                           6.7$ & $80\%$ &
                                         $18 \pm 2.8$ & $80\%$& $10.2
                                                                \pm
                                                                0.6$ &
                                                                       $80\%$ \\
    \hline
  \end{tabular}
  \caption{Number of steps taken to reach the goal in $7$D
    pick-and-place experiment for $5$ instances, each with random start
    and obstacle locations. We report mean
    and standard error \textit{only among}
    successful instances in which the robot reached the goal within $500$
    timesteps. The success subcolumn indicates percentage of
    successful instances.}
  \label{tab:pick-and-place}
\end{table*}

The task in this experiment is to pick and place a
heavy object from a shorter table, using a $7$ degree-of-freedom (DOF) robotic arm
(Figure~\ref{fig:intro} left) to a
goal pose on a taller table, while avoiding an obstacle.
As the object is heavy, the arm cannot generate the
required force in certain configurations and can only lift the object
to small heights. The problem is represented as planning in $7$D
discrete statespace where the first $6$ dimensions describe the $6$
DOF pose of the arm end-effector, and the last dimension
corresponds to the redundant DOF in the arm. The action
space $\actionspace$ is a discrete set of $14$ actions corresponding
to moving in each dimension by a fixed offset in the positive or
negative direction. The model $\Mhat$ used for planning models the
object as light, and hence does not capture the dynamics of the arm
correctly when it tries to lift the heavy object. The state space is
discretized into $10$ cells in each dimension resulting in a total of
$10^7$ states. Thus, we need to use
Algorithm~\ref{alg:large-state-spaces} for \cmaxpp{} and
\acmaxpp{}. The goal is to
pick and place the object for $20$ repetitions where at the start of
each repetition the object is in the start pose and needs to reach the
goal pose by the end of repetition.


We compare with \cmax{} for large state spaces, model-free
Q-learning~\cite{DBLP:conf/aaai/HasseltGS16}, and residual model
learning baselines~\cite{DBLP:conf/iros/SaverianoYFL17}. We chose two
kinds of function approximators for the learned residual dynamics:
global function approximators such as Neural Networks (NN) and local
memory-based function approximators such as K-Nearest Neighbors
regression (KNN.) Q-learning baseline uses $Q$-values that are
cleverly initialized using the model $\Mhat$ making it a strong model-free
baseline. We use the same neural network function
approximators for maintaining value estimates for all approaches and
perform $K=5$ expansions. We chose the metric $d$ as the manhattan
metric and use $\xi = 0$ for this experiment. We use a
radius of $\delta = 3$ for the hyperspheres introduced in the $7$D
discrete state space, and to ensure fair comparison use the same
radius for KNN regression. These values are chosen to reflect the
discrepancies observed when the arm tries to lift the object. All
approaches use the same initial value estimates obtained through
planning in $\Mhat$. \acmaxpp{} uses a non-increasing sequence
$\alpha_i = 1 + \beta_i$ where $\beta_1 = 4$ and $\beta_{i+1} =
0.5\beta_i$.

The results are presented in Table~\ref{tab:pick-and-place}. 
Model-free Q-learning takes a large number of executions in the
initial repetitions to estimate accurate $Q$-value estimates but in
later repetitions computes paths with lower costs managing to finish all
repetitions in $4$ out of $5$ instances. Among the residual model
learning baselines, the KNN approximator is successful in all
instances but takes a large number of executions to learn the true
dynamics, while the NN approximator finishes all repetitions in only
$2$ instances. \cmax{} performs well in the initial repetitions but
quickly gets stuck due to inflated costs and manages to complete the
task for $20$ repetitions in only $1$ instance. \cmaxpp{} is successful in
finishing the 
task in all instances and repetitions, while improving performance
across repetitions. Finally as expected, \acmaxpp{} also finishes all
repetitions, sometimes even having better performance than \cmax{} and
\cmaxpp{}.

\section{Discussion and Future Work}
\label{sec:disc-concl}

A major advantage of \cmaxpp{} is that, unlike previous approaches
that deal with inaccurate models, it can exploit inaccurately modeled
transitions without wasting online executions to learn the true
dynamics. It estimates the $Q$-value of incorrect transitions
leveraging past experience and enables the planner to compute solutions
containing such transitions. Thus, \cmaxpp{} is especially useful in
robotic domains with repetitive tasks where the true dynamics are
intractable to model, such as deformable manipulation, or vary
over time due to reasons such as wear and tear. Furthermore, the optimistic model assumption is
easier to satisfy, when compared to assumptions used by previous
approaches like \cmax{}, and performance of \cmaxpp{} degrades
gracefully with the accuracy of the model reducing to Q-learning in
the case where the model is inaccurate everywhere.
Limitations of
\cmaxpp{} and \acmaxpp{} include hyperparameters such as the
radius $\delta$ and the sequence $\{\alpha_i\}$, which might need to
be tuned for the task.
However, from our sensitivity experiments (see Appendix) we observe
that \acmaxpp{} performance is robust to the choice of sequence
$\{\alpha_i\}$ as long as it is non-increasing.
Note that Assumption~\ref{assumption:cmaxpp} can
be restrictive for tasks where designing an initial optimistic
model requires extensive domain
knowledge. However, it is infeasible to relax this assumption further
without resorting to global undirected exploration
techniques~\cite{Thrun-1992-15850}, which are highly sample
inefficient, to ensure completeness.

An interesting future direction is to interleave model identification
with \cmaxpp{} to combine the best of approaches that learn the true
dynamics and \cmaxpp{}. For instance, given a set of plausible forward
models we seek to quickly identify the best model while ensuring efficient
performance in each repetition.

\section*{Acknowledgements}
\label{sec:acknowledgements}

AV would like to thank Jacky Liang, Fahad Islam, Ankit Bhatia, Allie
Del Giorno, Dhruv Saxena and Pragna Mannam for their help in reviewing
the draft. AV is supported by the CMU presidential fellowship endowed
by TCS. Finally, AV would like to thank Caelan Garrett for developing
and maintaining the wonderful \texttt{ss-pybullet} library.

\bibliography{bib}

\appendix

\section{Sensitivity Experiments}
\label{sec:sens-exper}

In this section, we present the results of our sensitivity experiments
examining the performance of \acmaxpp{} with the choice of the
sequence $\{\alpha_i\}$. We compare the performance of different
choices of the sequence $\{\alpha_i\}$ on the $3$D mobile robot
navigation task. For each run, we average the results across $5$
instances with randomly placed ice patches and present the mean and
standard errors. To keep the figures concise, we plot the cumulative
number of steps taken to reach the goal from the start of the first
lap to the current lap across all laps. In all our runs, \acmaxpp{}
successfully completes all $200$ laps and hence, we do not report the
number of successful instances in our results.

We choose $4$ schedules for the sequence $\{\alpha_i\}$:
\begin{enumerate}
\item \textbf{Exponential Schedule}: In this schedule, we vary
  $\beta_{i+1} = \rho\beta_i$ where $\rho < 1$ is a constant that is
  tuned and $\alpha_i = 1 + \beta_i$. Observe that as $i\rightarrow
  \infty$, $\alpha_i \rightarrow 1$ and that the sequence
  $\{\alpha_i\}$ is a decreasing sequence.

  We vary both the initial $\beta_1$ chosen and the constant $\rho$ in
  our experiments. For $\beta_1$ we choose among values $[10, 100,
  1000]$ and $\rho$ is chosen among $[0.5, 0.7, 0.9]$. The results are
  shown in Figure~\ref{fig:exp}.
  \begin{figure}
  \centering
  \includegraphics[width=\linewidth]{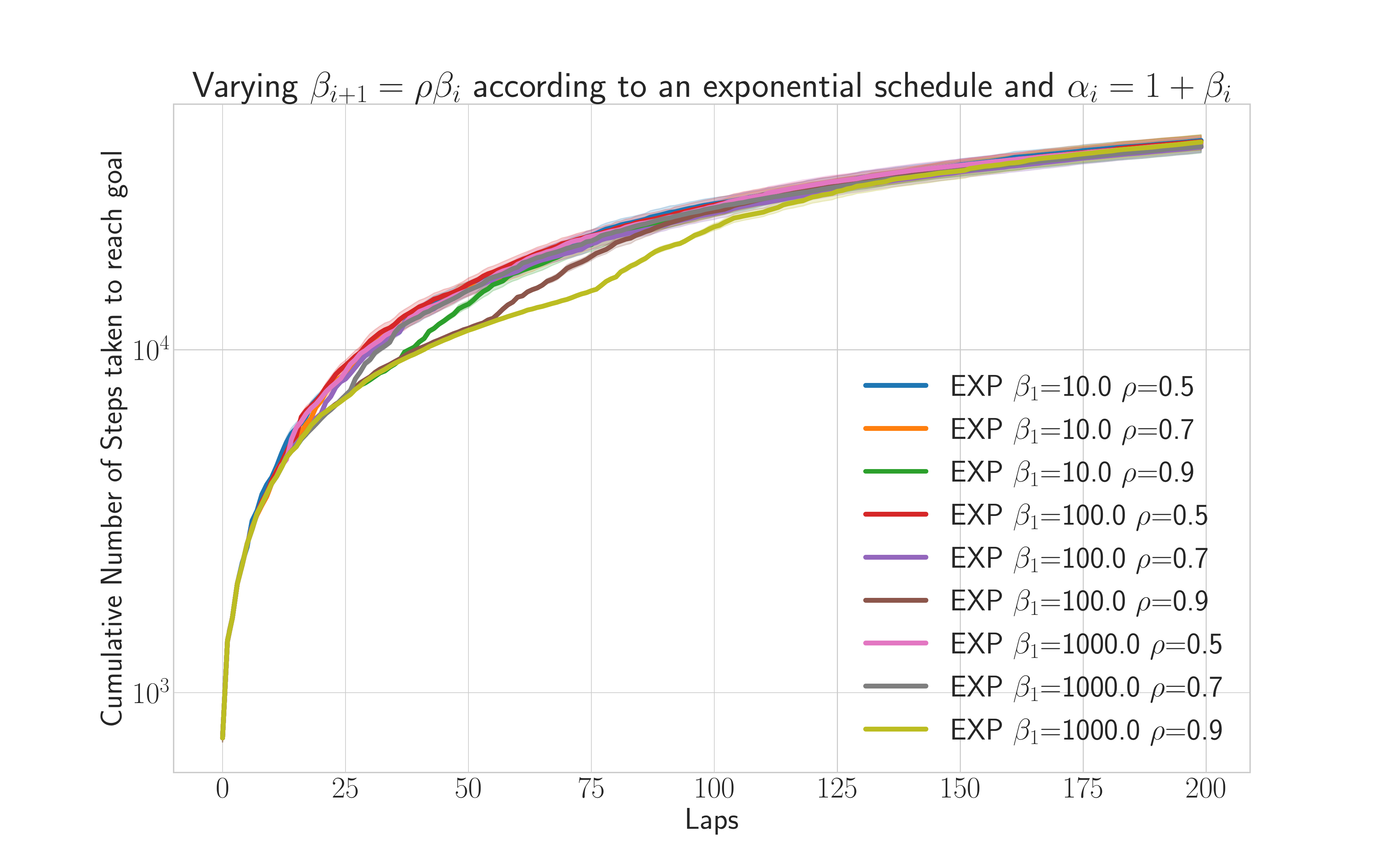}
  \caption{Sensitivity experiments with an exponential schedule}
  \label{fig:exp}  
\end{figure}

All choices have almost the same performance with $\beta_1 = 1000$ and
$\rho = 0.9$ having the best performance initially but has slightly
worse performance in the last several laps. The choice of $\beta_1 =
100$ and $\rho = 0.9$  seems to be a good choice with great performance in both
initial and final laps.

\item \textbf{Linear Schedule}: In this schedule, we vary $\beta_{i+1}
  = \beta_i - \eta$ where $\alpha_i = 1 + \beta_i$ and $\eta > 0$ is a
  constant that is determined
  so that $\beta_{200} = 0$, i.e. $\alpha_{200} = 1$. Hence, we have
  $\eta = \frac{\beta_1}{200}$.

  We vary the initial $\beta_1$ and choose among values $[10, 100,
  200]$. The results are shown in Figure~\ref{fig:linear}.
  \begin{figure}
  \centering
  \includegraphics[width=\linewidth]{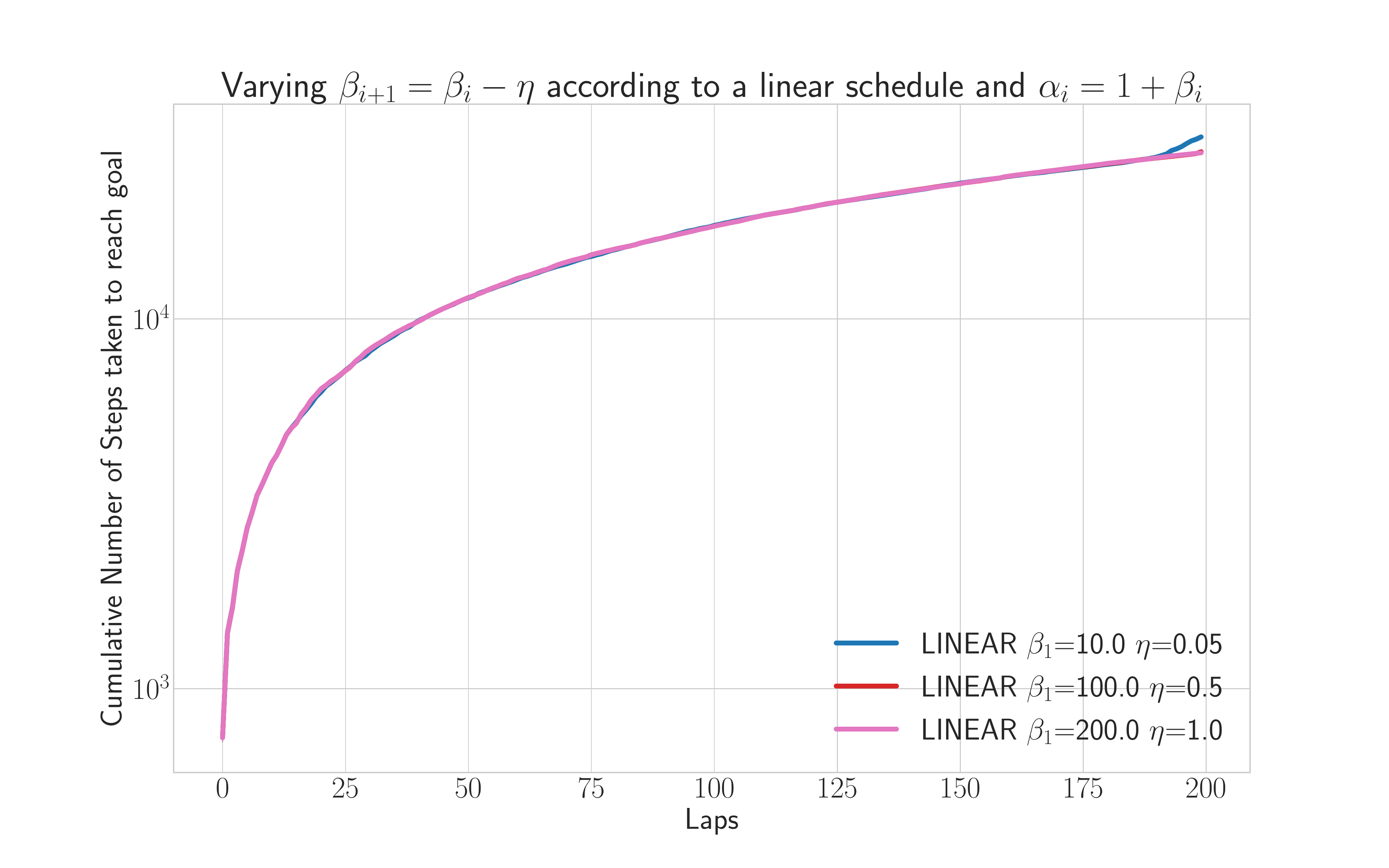}
  \caption{Sensitivity experiments with a linear schedule}
  \label{fig:linear}
\end{figure}

All three choices have the same performance except in the last few
laps where $\beta_1 = 10$ degrades while the other two choices perform well.
\item \textbf{Time Decay Schedule}: In this schedule, we vary
  $\beta_{i+1} = \frac{\beta_1}{i+1}$ where $\alpha_i = 1 +
  \beta_i$. In other words, we decay $\beta$ at the rate of
  $\frac{1}{i}$ where $i$ is the lap number. Again, observe that as $i
  \rightarrow \infty$, we have $\alpha_i \rightarrow 1$.

  We vary the initial $\beta_1$ and choose among values $[10, 100,
  1000]$. The results are shown in Figure~\ref{fig:time}.
  \begin{figure}
  \centering
  \includegraphics[width=\linewidth]{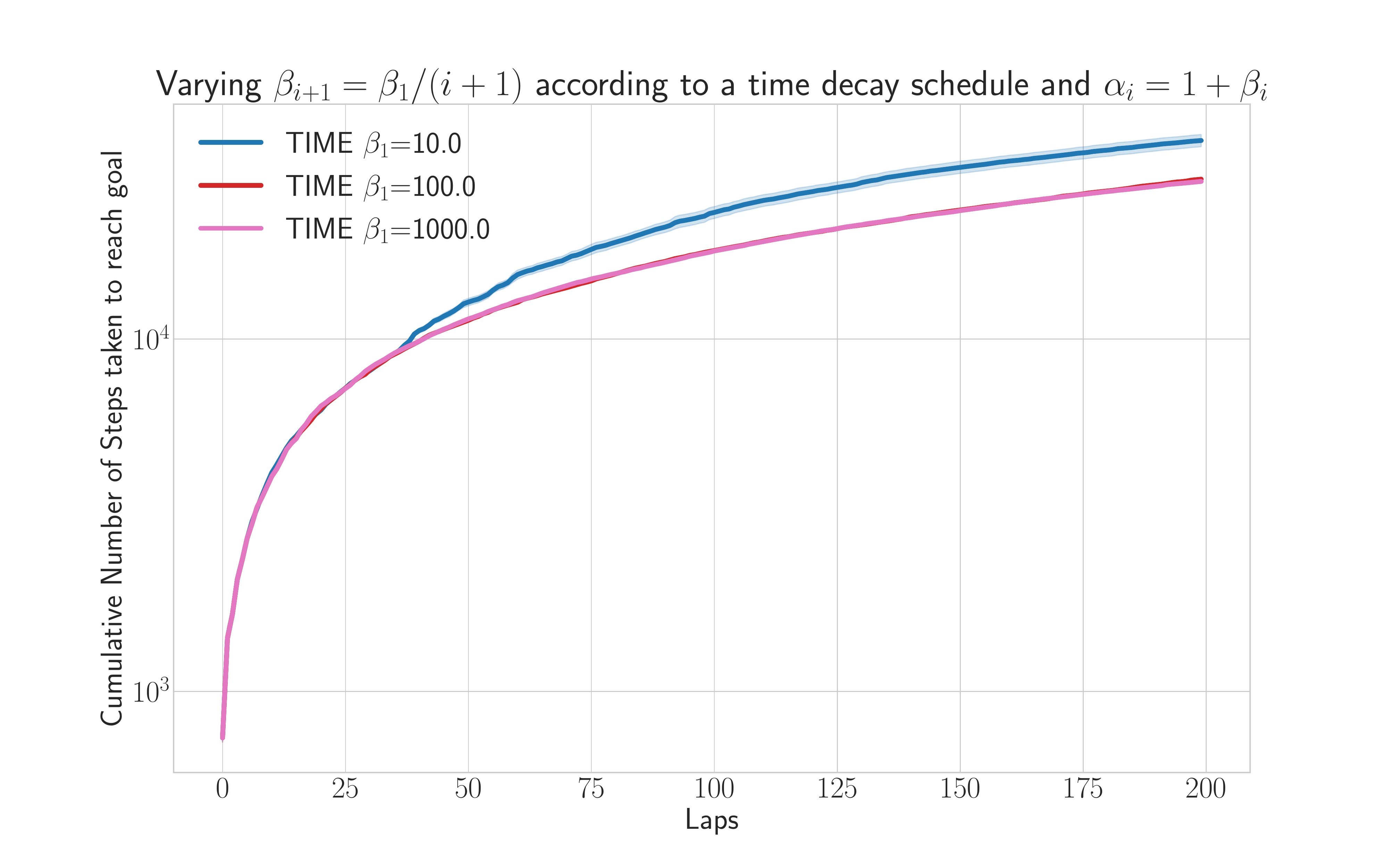}
  \caption{Sensitivity experiments with a time decay schedule}
  \label{fig:time}
\end{figure}

The choices of $\beta_1 = 100$  and $\beta_1 = 1000$ have the best
(and similar) performance while $\beta_1 = 10$ has a poor performance
as it quickly switches to \cmaxpp{} in the early laps and wastes
executions learning accurate $Q$-values.
\item \textbf{Step Schedule}: In this schedule, we vary $\beta$ as a
  step function with $\beta_{i+1} = \beta_i - \delta$ if $i$ is a
  multiple of $\xi$ where $\xi$ is the step frequency, $\alpha_i = 1 +
  \beta_i$ and $\delta$ is a constant that is determined so that
  $\beta_{200} = 0$, i.e. $\alpha_{200} = 1$. Hence, we have $\delta =
  \frac{\beta_1\xi}{200}$.

  We vary both the initial $\beta_1$ and the step frequency $\xi$. For
  $\beta_1$ we choose among values $[10, 100, 200]$ and for $\xi$ we
  choose among $[5, 10, 20]$. The results are shown in
  Figure~\ref{fig:step}.
  \begin{figure}
  \centering
  \includegraphics[width=\linewidth]{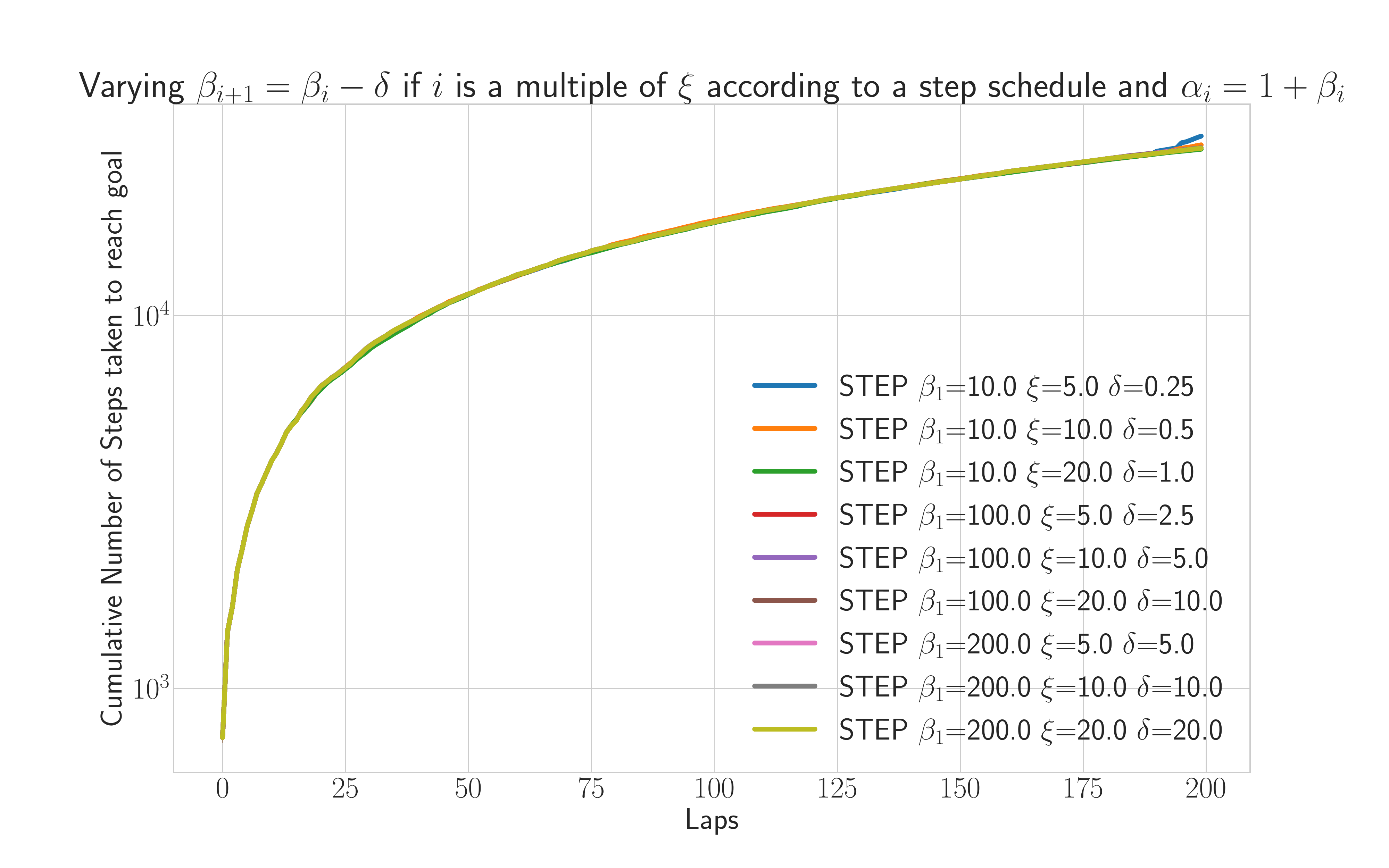}
  \caption{Sensitivity experiments with a step schedule}
  \label{fig:step}
\end{figure}

All choices have the same performance and \acmaxpp{} seems to be
robust to the choice of step size frequency.
\end{enumerate}

\begin{figure}
  \centering
  \includegraphics[width=\linewidth]{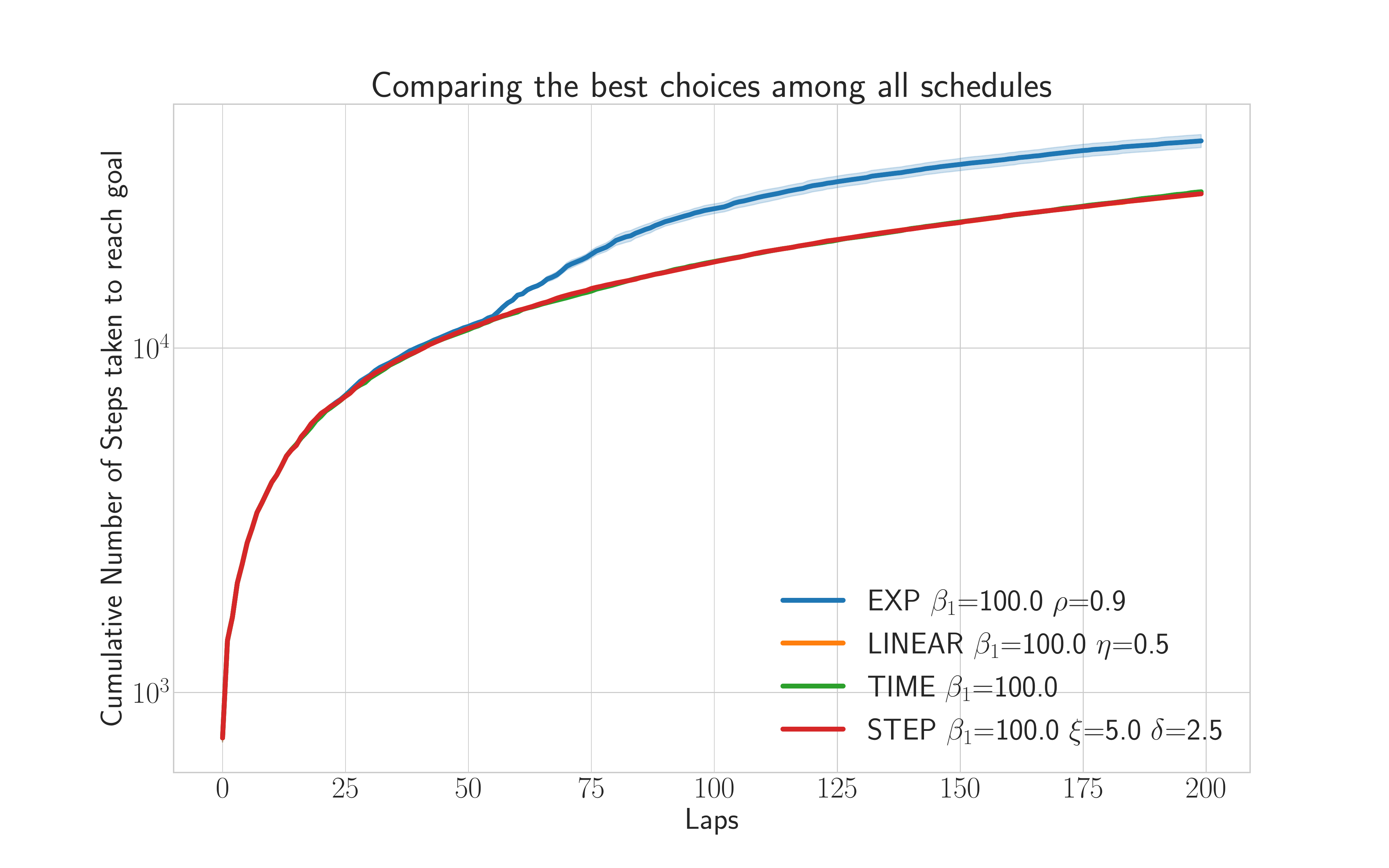}
  \caption{Sensitivity experiments with best choices among all
    schedules}
  \label{fig:best}
\end{figure}

For our final comparison, we will pick the best performing choice
among all the schedules and compare performance among these selected
choices. The results are shown in Figure~\ref{fig:best}.

We can observe that all schedules have the same performance except the
exponential schedule which has worse performance. This can be
attributed to the rapid decrease in the value of $\beta$ compared to
other schedules and thus, around lap $50$ \acmaxpp{} switches to
\cmaxpp{} resulting in a large number of executions wasted to learn
accurate $Q$-value estimates. This does not happen for other schedules
as they decrease $\beta$ gradually and thus, spreading out the
executions used to learn accurate $Q$-value estimates across several
laps and not performing poorly in any single lap.

\section{Proofs}
\label{sec:proofs}

In this section, we present the assumptions and proofs that result in
the theoretical guarantees of \cmaxpp{} and \acmaxpp{}.

\subsection{Significance of Optimistic Model Assumption}
\label{sec:sign-optim-model}

To understand the significance of the optimistic model assumption, it
is important to note that completeness guarantees usually require the
use of admissible and consistent value estimates, i.e. estimates that
always underestimate the true cost-to-goal values. This requirement
needs to hold every time we plan (or replan) to ensure that we never
discard a path as being expensive in terms of cost, when it is cheap
in reality.

All our guarantees assume that the initial value estimates are
consistent and admissible, but to ensure that they always remain
consistent and admissible throughout execution, we need the optimistic
model assumption. This assumption ensures that updating value
estimates by planning in the model $\Mhat$ always results in estimates
that are admissible and consistent. In other words, the optimal value
function (which we obtain by doing full state space planning in
$\Mhat$) of the model $\Mhat$ always underestimates the optimal value
function of the environment $M$ at all states $s \in \statespace$.

A very intuitive way to understand the assumption is to imagine a
navigation task where the robot is navigating from a start to goal in
the presence of obstacles. In this example, the optimistic model
assumption requires that the model should never place an obstacle in a
location when there isn't an obstacle in the environment at the same
location. However, if there truly is an obstacle at some location,
then the model can either have an obstacle or not have one at the same
location. Put simply, an agent that is planning using the model should
never be ``pleasantly surprised" by what it sees in the
environment. Several other intuitive examples are presented
in~\cite{DBLP:conf/aaai/Jiang18} and we recommend the reader to look
at them for more intuition.

\subsection{Completeness Proof}
\label{sec:completeness-proof}

To prove completeness, first we need to note that the $Q$-update in
\cmaxpp{} always ensures that the $Q$-value estimates remain
consistent and admissible as long as the state value estimates remain
consistent and admissible. We have already seen why the optimistic
model assumption ensures that the state value estimates always remain
consistent and admissible. Thus, we can use Theorem~3 from
RTAA*~\cite{DBLP:conf/atal/KoenigL06} in conjunction with the
optimistic model assumption to ensure completeness. Note that if the
model is inaccurate everywhere, then our planner reduces to doing
$K=1$ expansions at every time step and acts similar to $Q$-learning,
which is also guaranteed to be complete with admissible and consistent
estimates~\cite{DBLP:conf/aaai/KoenigS93}. The worst case bound of
$|\statespace|^3$ steps is taken directly from the upper bound on
$Q$-learning from~\cite{DBLP:conf/aaai/KoenigS93}. The above arguments
are true for both \cmaxpp{} and \acmaxpp{}. Note that for \acmaxpp{}
if all paths to the goal contains an incorrect transition then the
penalized value estimate $\Vtilde(s) > \alpha V(s)$ for any finite
$\alpha$ and thus, will fall back on \cmaxpp{}.

For the second part of the theorem, the assumption of \cmax{} (we will
refer this as \textit{optimistic penalized model assumption}) in
conjunction with RTAA* guarantee again ensures completeness for
\cmaxpp{} and \acmaxpp{}. To see this, observe that the optimistic
penalized model assumption ensures that the value estimates are always
admissible and consistent w.r.t the true penalized model
($\Mtilde_\incorrectset$ where $\incorrectset$ contains all the
incorrect transitions) and from the assumption, we know that there
exists a path to the goal in the true penalized model. Hence,
\cmaxpp{} and \acmaxpp{} are bound to find this path.

\cmaxpp{} again utilizes the worst case bounds of $Q$-learning under
the optimistic penalized assumption as well and attains an upper bound
of $|\statespace|^3$ steps. However, \acmaxpp{} with a sufficiently
large $\alpha_i$ for any repetition $i$ acts similar to \cmax{}, and
thereby can utilize the worst case bounds of LRTA* (which is simply
RTAA* with $K=1$ expansions) from~\cite{DBLP:conf/aaai/KoenigS93}
giving an upper bound of $|\statespace|^2$ time steps. This shows the
advantage of \acmaxpp{} over \cmaxpp{}, especially in earlier
repetitions when the incorrect set $\incorrectset$ is small (thus,
making the optimistic penalized model assumption hold,) and $\alpha_i$
is large.

\subsection{Asymptotic Convergence Proof}
\label{sec:asympt-conv-proof}

The asymptotic convergence proof completely relies on the asymptotic
convergence of $Q$-learning~\cite{DBLP:conf/aaai/KoenigS93} and
asymptotic convergence of LRTA*~\cite{DBLP:journals/ai/Korf90} to
optimal value estimates. The proof again crucially relies on the fact
that the value estimates always remain admissible and consistent,
which is ensured by the optimistic model assumption. Note that the
optimistic penalized model assumption is not enough to guarantee
asymptotic convergence to the optimal cost in $M$ as we penalize
incorrect transitions. However, it is possible to show that under the
optimistic penalized model assumption both \cmaxpp{} and \acmaxpp{}
converge to the optimal cost in the true penalized model
$\Mtilde_\incorrectset$ where $\incorrectset$ contains all incorrect
transitions.

\section{Experiment Details}
\label{sec:experiment-details}

All experiments were implemented using Python 3.6 and run on a
$3.1$GHz Intel Core i$5$ machine. We use
PyTorch~\cite{NEURIPS2019_9015} to train neural network function
approximators in our $7$D experiments, and use Box2D~\cite{catto2007box2d} for our 3D mobile
robot simulation (similar to OpenAI Gym~\cite{brockman2016openai}
\texttt{car\_racing} environment) and use
PyBullet~\cite{coumans2013bullet} for our $7$D PR2 experiments.

\subsection{3D Mobile Robot Navigation with Icy Patches}
\label{sec:3d-mobile-robot}

\begin{figure}[t]
  \centering
  \includegraphics[width=\linewidth]{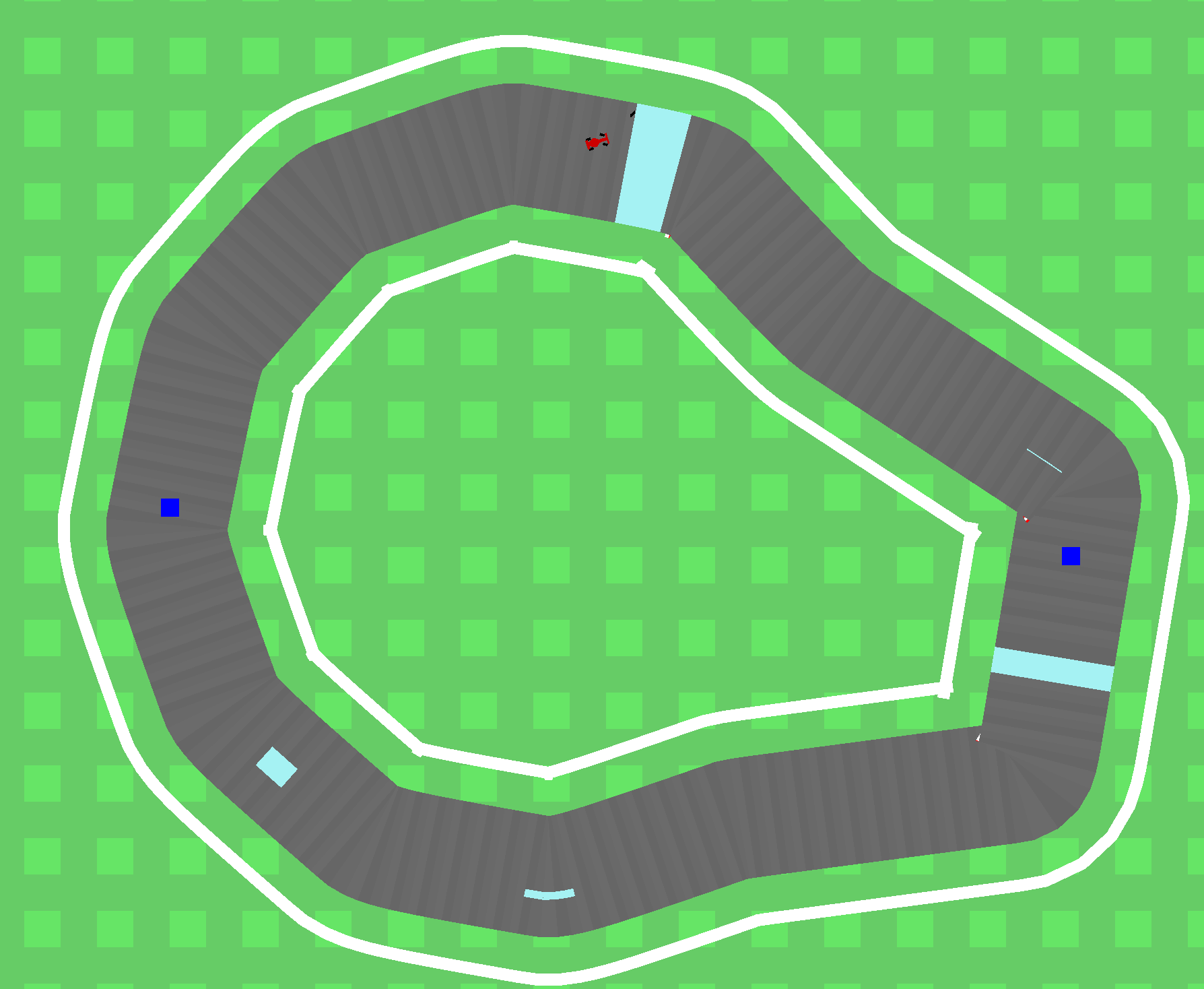}
  \caption{$3$D Mobile Robot experiment example track}
  \label{fig:track}
\end{figure}

An example track used in the $3$D experiment is shown in
Figure~\ref{fig:track}. We generate $66$ motion primitives offline
using the following procedure: (a) We first define the primitive
action set for the robot by discretizing the steering angle into 3
cells, one corresponding to zero and the other two corresponding to 
$+0.6$ and $-0.6$ radians. We also discretize the speed of the robot
to 2 cells corresponding to $+2$m/s and $-2$m/s, (b) We then
discretize the state space into a $100\times 100$ grid in $XY$ space
and $16$ cells in $\theta$ dimension. Thus, we have a $100 \times 100
\times 16$ grid in $XY\theta$ space., (c) We then initialize the robot
at $(0, 0)$ $xy$ location with different headings chosen among $[0,
\cdots, 15]$ and roll out all possible sequences of primitive actions
for all possible motion primitive lengths from $1$ to $15$ time steps,
(d) We filter out all motion primitives whose end point is very close
to a cell center in the $XY\theta$ grid. During execution, we use a
pure pursuit controller to track the motion primitive so that the robot
always starts and ends on a cell center. During planning, we simply
use the discrete offsets stored in the motion primitive to compute the
next state (and thus, the model dynamics are pre-computed offline
during motion primitive generation.)

The cost function used is as follows: for any motion primitive $a$ and
state $s$,
the cost of executing $a$ from $s$ is given by $c(s, a) = \sum_{s'}
c'(s')$ where $c'$ is a pre-defined cost map over the $100 \times 100
\times 16$ grid and $s'$ is all the intermediate states (including the
final state) that the robot
goes through while executing the motion primitive $a$  from $s$. The
pre-defined cost map is defined as follows: $c'(s) = 1$ if state $s$
lies on the track (i.e. $xy$ location corresponding to $s$ lies on the
track) and $c'(s) = 100$ otherwise (i.e. all $xy$ locations
corresponding to grass or wall has a cost of $100$). This encourages
the planner to come up with a path that lies completely on the track.

We define two checkpoints on the opposite ends of the track (shown as
blue squares in Figure~\ref{fig:track}.) The goal of the robot is to
reach the next checkpoint incurring least cost while staying on the
track. Note that this 
requires the robot to complete laps around the track as quickly as
possible. Since the state space is small, we maintain value estimates
$V, Q, \tilde{V}$ using tables and update the appropriate table entry
for each value update. The tables are initialized with value estimates
obtained by planning in the model $\Mhat$ using a planner with $K=100$
expansions until the robot can efficiently complete the laps using the
optimal paths. However, this does not mean that the initial value
estimates are the optimal values for $\Mhat$ dynamics since the
planner looks ahead and can achieve optimal paths with underestimated
value functions. Nevertheless, these estimates are highly informative.

\subsection{7D Pick-and-Place with a Heavy Object}
\label{sec:7d-pick-place}

\begin{figure}[t]
  \centering
  \includegraphics[width=\linewidth]{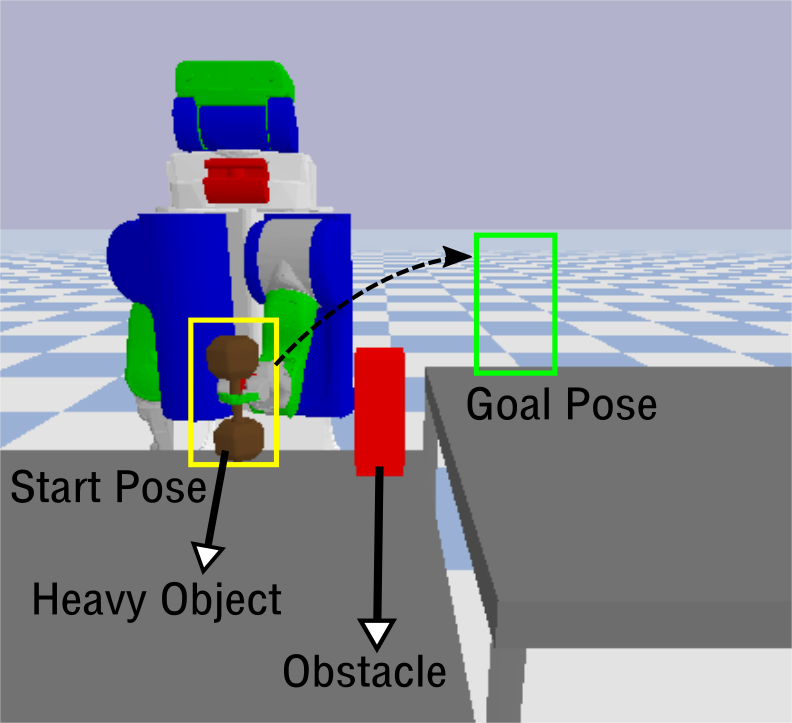}
  \caption{$7$D Pick-and-Place Experiment}
  \label{fig:pr2}
\end{figure}

For our $7$D experiments, we make use of Bullet Physics Engine through
the pyBullet interface. For motion planning and other simulation
capabilities we make use of \texttt{ss-pybullet}
library~\cite{sspybullet}. The task is shown in
Figure~\ref{fig:pr2}. The goal is for the robot to pick the heavy
object from its start pose and place it at its goal pose while
avoiding the obstacle, without any resets. Since the object is heavy,
the robot fails to lift the object in certain configurations where it
cannot generate the required torque to lift the object. Thus, the
robot while lifting the object might fail to reach the goal waypoint
and onky reach an intermediate waypoint resulting in discrepancies
between modeled and true dynamics.

This is represented as a planning problem in $7$D statespace. The
first $6$ dimensions correspond to the $6$DOF pose of the object (or
gripper,) and the last dimension corresponds to the redundant DOF in
the arm (in our case, it is the upper arm roll joint.) Given a $7$D
configuration, we use IKFast library~\cite{diankov_thesis} to compute
the corresponding $7$D joint angle configuration. The action space
consists of $14$ motion primitives that move the arm by a fixed offset
in each of the $7$ dimensions in positive and negative directions. The
discrepancies in this experiment are only in the $Z$ dimension
corresponding to lifting the object. For planning, we simply use a
kinematic model of the arm and assume that the object being lifted is
extremely light. Thus, we do not need to explicitly consider dynamics
during planning. However, during execution we take the dynamics into
account by executing the motion primitives in the simulator. The cost
of any transition is $1$ if the object is not at goal pose, $0$ if the
object is at goal pose. We start the next repetition only if the robot
reached the goal pose in the previous repetition.

The $7$D state space is discretized into $10$ cells in each dimension
resulting in $10^7$ states. Since the state space is large we use
neural network function approximators to maintain the value functions
$V, Q, \tilde{V}$. For the state value functions $V, \tilde{V}$ we use
the following neural network approximator: a feedforward network with
$3$ hidden layers consisting of $64$ units each, we use ReLU
activations after each layer except the last layer, the network takes
as input a $34$D feature representation of the $7$D state computed as
follows:

\begin{itemize}
	\item For any discrete state $s$, we compute a continuous
          $10$D representation $r(s)$ that is used to construct the
          features
          
	\begin{itemize}
		\item The discrete state is represented as $(xd, yd,
                  zd, rd, pd, yd, rjointd)$ where $(xd, yd, zd)$
                  represents the $3$D discrete location of the object
                  (or gripper,) $(rd, pd, yd)$ represents the discrete
                  roll, pitch, yaw of the object (or gripper,) and
                  $rjointd$ represents the discrete redundant joint
                  angle
                  
		\item We convert $(xd, yd, zd)$ to a continuous
                  representation by simply dividing by the grid size
                  in those dimensions, i.e. $(xc, yc, zc) = (xd/10,
                  yd/10, zd/10)$
                  
		\item We do a similar construction for $rjointc$,
                  i.e. $rjointc = rjointd/10$
                  
		\item However, note that $rd, pd, yd$ are angular
                  dimensions and simply dividing by grid size would
                  not encode the wrap around nature that is inherent
                  in angular dimensions (we did not have this problem
                  for $rjointd$ as the redundant joint angle has lower
                  and upper limits, and is always recorded as a value
                  between those limits.) To account for this, we use a
                  sine-cosine representation defined as $(rc1, rc2,
                  pc1, pc2. yc1, yc2) = (sin(rc), cos(rc), sin(pc),
                  cos(pc), sin(yc), cos(yc))$ where $rc, pc, yc$ are
                  the roll, pitch, yaw angles corresponding to the
                  cell centers of the grid cells $rd, pd, yd$.
                  
		\item Thus, the final $10$D representation of state
                  $s$ is given by $r(s) = (xc, yc, zc, rc1, rc2, pc1,
                  pc2, yc1, yc2, rjointc)$
                  
		\item We also define a truncated $9$D representation
                  $r'(s) = (xc, yc, zc, rc1, rc2, pc1, pc2, yc1, yc2)$
                  and a $3$D representation $r''(s) = (xc, yc, zc)$
                  
	\end{itemize}
	\item The first feature is the $9$D relative position of the $6$D goal pose w.r.t the object $f1 = r'(g) - r'(s)$
	\item The second feature is the $10$D relative position of the object w.r.t the gripper home state $h$, $f2 = r(s) - r(h)$
	\item The third feature is the $9$D relative position of the goal w.r.t the gripper home state $h$, $f3 = r'(g) - r'(h)$
	\item The fourth feature is the $3$D relative position of the obstacle left top corner $o1$ w.r.t the object, $f4 = r''(o1) - r''(s)$
	\item The fifth and final feature is the $3$D relative position of the object right bottom corner $o2$ w.r.t. the object, $f5 = r''(o2) - r''(s)$
	\item Thus, the final $34$D feature representation is given by $f(s) = (f1, f2, f3, f4, f5)$.
\end{itemize}

The output of the network is a single scalar value representing the
cost-to-goal of the input state. Instead of learning the
cost-to-goal/value from scratch, we start with an initial value
estimate that is hardcoded (manhattan distance to goal in the $7$D
discrete grid) and the neural network approximator is used to learn a
residual on top of it. A similar trick was used in
\cmax{}~\cite{Vemula-RSS-20}. The residual state value function
approximator was initialized to output $0$ for all $s \in
\statespace$. We use a similar architecture for the residual $Q$-value
function approximator but it takes as input the $34$D state feature
representation and outputs a vector in $\mathbb{R}^{|\actionspace|}$
(in our case, $\mathbb{R}^{14}$) to represent the cost-to-goal
estimate for each action $a \in \actionspace$. We also use the same
hardcoded value estimates as before in addition to the residual
approximator to construct the $Q$-values. All baselines and proposed
approaches use the same function approximator and same initial
hardcoded value estimates to ensure fair comparison. The value
function approximators are trained using mean squared loss.

The residual model learning baseline with neural network (NN) function
approximator uses the following architecture: $2$ hidden layers each
with $64$ units and all layers are followed by ReLU activations except
the last layer. The input of the network is the $34$D feature
representation of the state and a one-hot encoding of the action in
$\mathbb{R}^14$. The output of the network is the $7$D continuous
state which is added to the state predicted by the model $\Mhat$. The
loss function used to train the network is a simple mean squared
loss. The residual model learning baseline with K-Nearest Neighbor
regression approximator (KNN) uses a manhattan radius of $3$ in the
discrete $7$D state space. We compute the prediction by averaging the
next state residual vector observed in the past for any state that
lies within the radius of the current state. The averaged residual is
added to the next state predicted by model $\Mhat$ to obtain the
learned next state.

We use Adam optimizer~\cite{DBLP:journals/corr/KingmaB14} with a
learning rate of $0.001$ and a weight decay (L$2$ regularization
coefficient) of $0.001$ to train all the neural network function
approximators in all approaches. We use a batch size of $32$ for the
state value function approximators and a batch size of $128$ for the
$Q$-value function approximators. We perform $U = 3$ updates for state
value function and $U = 5$ updates for state-action value function for
each time step. We update the parameters of all neural network
approximators using a polyak averaging coefficient of $0.5$.

Finally, we use hindsight experience replay
trick~\cite{DBLP:conf/nips/AndrychowiczCRS17} in training all the
value function approximators with the probability of sampling any
future state in past trajectories as the goal set to $0.7$. This is
crucial as our cost function used is extremely sparse.

\end{document}